\documentclass[conference]{IEEEtran}
\IEEEoverridecommandlockouts
\usepackage{cite}
\usepackage{amsmath,amssymb,amsfonts}
\usepackage{algorithmic}
\usepackage{graphicx}
\usepackage{textcomp}
\usepackage{xcolor}
\usepackage{multicol}
\usepackage{multirow}
\usepackage{amsmath}
\usepackage{url}

\def\BibTeX{{\rm B\kern-.05em{\sc i\kern-.025em b}\kern-.08em
    T\kern-.1667em\lower.7ex\hbox{E}\kern-.125emX}}
\begin{document}

\title{A Biologically Inspired Feature Enhancement Framework for Zero-Shot Learning\\
\thanks{Submit to IEEE CSCloud 2020 Special Session on Artificial Intelligence and Security (AIS). This work was supported by the Opening Project of Shanghai Trusted Industrial Control Platform (TICPSH202003008-ZC), National Natural Science Foundation of China (61672358, 61836005, 61976141, 61732011), Guangdong Science and Technology Department (Grant no. 2018B010107004), and in part by Basic Research Project of Knowledge Innovation Program in Shenzhen (Grant nos. JCYJ20180305125850156) (*Corresponding author)}
}
\author{\IEEEauthorblockN{1\textsuperscript{st} Zhongwu Xie}
\IEEEauthorblockA{\textit{College of Computer Science and Software Engineering} \\
\textit{Shenzhen University}\\
Shenzhen, China \\
1810272010@email.szu.edu.cn}
\and
\IEEEauthorblockN{2\textsuperscript{nd} Weipeng Cao*}
\IEEEauthorblockA{\textit{College of Computer Science and Software Engineering} \\
\textit{Shenzhen University}\\
Shenzhen, China \\
caoweipeng@szu.edu.cn}
\and
\IEEEauthorblockN{3\textsuperscript{rd} Xizhao Wang}
\IEEEauthorblockA{\textit{College of Computer Science and Software Engineering} \\
\textit{Shenzhen University}\\
Shenzhen, China \\
xizhaowang@ieee.org}
\and
\IEEEauthorblockN{4\textsuperscript{th} Zhong Ming}
\IEEEauthorblockA{\textit{College of Computer Science and Software Engineering} \\
\textit{Shenzhen University}\\
Shenzhen, China \\
mingz@szu.edu.cn}
\and
\IEEEauthorblockN{5\textsuperscript{th} Jingjing Zhang*}
\IEEEauthorblockA{\textit{College of Command and Control Engineering} \\
\textit{Army Engineering University of People's Liberation Army of China}\\
Nanjing, China \\
lg02103@163.com}
\and
\IEEEauthorblockN{6\textsuperscript{th} Jiyong Zhang}
\IEEEauthorblockA{\textit{School of Automation} \\
\textit{Hangzhou Dianzi University}\\
Hangzhou, China \\
jzhang@hdu.edu.cn}
}
\maketitle

\begin{abstract}
Most of the Zero-Shot Learning (ZSL) algorithms currently use pre-trained models as their feature extractors, which are usually trained on the ImageNet data set by using deep neural networks. The richness of the feature information embedded in the pre-trained models can help the ZSL model extract more useful features from its limited training samples. However, sometimes the difference between the training data set of the current ZSL task and the ImageNet data set is too large, which may lead to the use of pre-trained models has no obvious help or even negative impact on the performance of the ZSL model. To solve this problem, this paper proposes a biologically inspired feature enhancement framework for ZSL. Specifically, we design a dual-channel learning framework that uses auxiliary data sets to enhance the feature extractor of the ZSL model and propose a novel method to guide the selection of the auxiliary data sets based on the knowledge of biological taxonomy. Extensive experimental results show that our proposed method can effectively improve the generalization ability of the ZSL model and achieve state-of-the-art results on three benchmark ZSL tasks. We also explained the experimental phenomena through the way of feature visualization.

\end{abstract}


\section{Introduction}
Machine learning models, especially neural networks, have shown great potential in many fields in recent years~\cite{cao2018review,cao2018fuzziness,cao2017fuzziness}. Most of these existing models follow an implicit assumption, that is, the classes of the testing samples must exist in the classes set of the training samples. However, in real-life applications, sometimes the classes of testing samples never appear in the training data set. In this case, we call the classes of these testing samples unseen classes, which corresponds to the classes that have appeared in the training data set (i.e., the seen classes). Traditional machine learning algorithms are often unable to accurately predict the classes of the samples belonging to the unseen classes. Zero-Shot Learning (ZSL) is one of the effective algorithms proposed to solve this problem \cite{lampert2009learning,palatucci2009zero,akata2013label,xian2017zero,luo2020novel} .

In ZSL, both the seen and unseen classes are described by specific semantic vectors in the semantic space (i.e., the side information). The commonly used semantic vectors include the attribute \cite{farhadi2009describing}, word2vec \cite{socher2013zero}, sentences \cite{reed2016learning}, and gaze \cite{karessli2017gaze}. For a specific data set, this side information is encoded into vectors with the same dimension, and each class corresponds to a specific vector. ZSL establishes the mapping relationship between the visual space where the images' feature vectors of the seen classes are located and the semantic space where the side information is located to obtain the learner with strong generalization ability and then applies it to predict the classes of the testing samples. In other words, ZSL directly builds the internal relationship between the visual space and the semantic space based on the training samples of the seen classes and then extends it to predict the labels of the testing samples.

According to the different prediction targets of ZSL, the current ZSL algorithms can be grouped into two categories: Conventional Zero-Shot Learning (CZSL) \cite{lampert2009learning} and Generalized Zero-Shot Learning (GZSL) \cite{xian2018zero}. They differ in that the testing samples of CZSL come only from the unseen classes, while the testing samples of GZSL can come from both the seen and unseen classes. Compared with CZSL, GZSL is more consistent with practical applications, because the samples tested may come from both the seen and unseen classes in reality. However, GZSL suffers from a serious bias problem \cite{xian2018feature,zhu2018generative}, that is, since the ZSL model has never seen samples of the unseen classes, it tends to predict the labels of these samples as the classes of the seen classes similar to their real labels. Although ZSL has great potential advantages in solving real-world tasks, the research on ZSL is still in its infancy due to the difficulty of the problem, many fundamental problems have not been effectively solved, such as the quality of feature extraction in ZSL cannot be guaranteed at present.

Specifically, most of the current ZSL algorithms use pre-trained models, which are usually trained on the ImageNet data set \cite{deng2009imagenet}, to transfer the training data set of the ZSL tasks into feature vectors, and then focus on building the mapping relationship between the feature space and the semantic space. This method improves the feature extraction ability of the ZSL model on the ZSL training data set with the help of the rich feature information embedded in the pre-trained models. However, intuitively, if the difference between the training data set of the current ZSL task and the training data set used by the pre-trained model is too large, this method may not work well. For example, given two pre-trained models $M1$ and $M2$, suppose $M1$ is trained with the data set containing only fruit images, while $M2$ is trained with the data set containing only cat images. If the current ZSL task is to classify the dog species, then intuitively using $M2$ may be more beneficial to the accurate classification of the final ZSL model. In other words, we think that if there is a strong correlation between the data set used by the pre-trained model and the training data set of the current ZSL task, then the pre-trained model may have a positive impact on the predictive ability of the ZSL model and vice versa. However, it is too expensive to collect and label a relevant large data set like ImageNet to train a specific pre-trained model. The best compromise is to fine-tune the existing pre-trained models to fit the current task.

To solve this problem, based on the idea of multi-task learning, we design a dual-channel learning framework to enhance the feature extraction ability of the pre-trained model used in ZSL by using auxiliary data sets. Specifically, we choose some image samples from ImageNet that are most relevant to the seen classes of the current ZSL task to form the auxiliary data set and then put it and the original ZSL training data set into our proposed framework to train the model. The auxiliary data set can regularize the feature extractor of ZSL (i.e., the pre-trained model) and make it provide more relevant features for the current task.

But how to choose the relevant auxiliary data set? In other words, how to measure the correlation strength between an image data set and the current image classification task? It is still very difficult to measure the similarity of two image data sets mathematically. However, this problem is not difficult for human beings, even for children. Because biologists have built a relatively complete body of knowledge to distinguish the relationship between two species.

Inspired by this observation, we propose a novel biological taxonomy-based data set selection method to help us to select the auxiliary data set. Specifically, biologists currently divide the degree of kinship of all things in the world into seven levels: $Kingdom$, $Phylum$, $Class$, $Order$, $Family$, $Genus$, and $Species$. $Species$ is the most basic and specific taxonomic rank. In other words, the same $species$ means the strongest correlation. With the gradual expansion of the taxonomic rank (i.e., $genus, family, order, class, phylum, kingdom$), the degree of correlation gradually decreases. We use this biological knowledge system to guide the selection of the auxiliary data set.

To our best knowledge, we are the first to design the dual-channel learning framework for ZSL to enhance its feature extractor and introduce the biological correlation to select the auxiliary data set. The main contributions of this study are summarized as follows.

(1) Based on the idea of multi-task learning, we propose a dual-channel learning framework for ZSL, which can enhance the feature extractor of ZSL with the help of the auxiliary data set and improve the generalization ability of the ZSL model.

(2) We propose a novel auxiliary data set selection strategy based on the knowledge of biological taxonomy, which can effectively measure the correlation degree between image data sets.

(3) Our study found that under the learning framework we proposed, the performance of the ZSL model shows a linear increasing trend with the increasing degree of the correlation between the auxiliary data set and the current task, which implies that the performance of the ZSL model can be greatly improved when the feature extractor is fine-tuned by using the most relevant auxiliary tasks. The findings promise to provide a new way of thinking for a follow-up study of the ZSL algorithm.

(4) The experimental results on three benchmark datasets show that our proposed method can effectively improve the generalization ability of the ZSL model and achieve state-of-the-art results on three benchmark ZSL tasks. Moreover, we use the method of feature visualization to explain the experimental phenomena.

The remainder of this paper is organized as follows. In Sec. II, we introduce the necessary preliminaries including the general knowledge of biological taxonomy, multi-task learning, and zero-shot learning. We give the details of our proposed method in Sec. III. The experimental results and the corresponding analysis are given in Sec. IV. In Sec. V, we conclude this paper.

\section{Preliminaries}

\subsection{General knowledge of biological taxonomy}
\begin{figure}\centering
    \includegraphics[width=8cm,height=5cm]{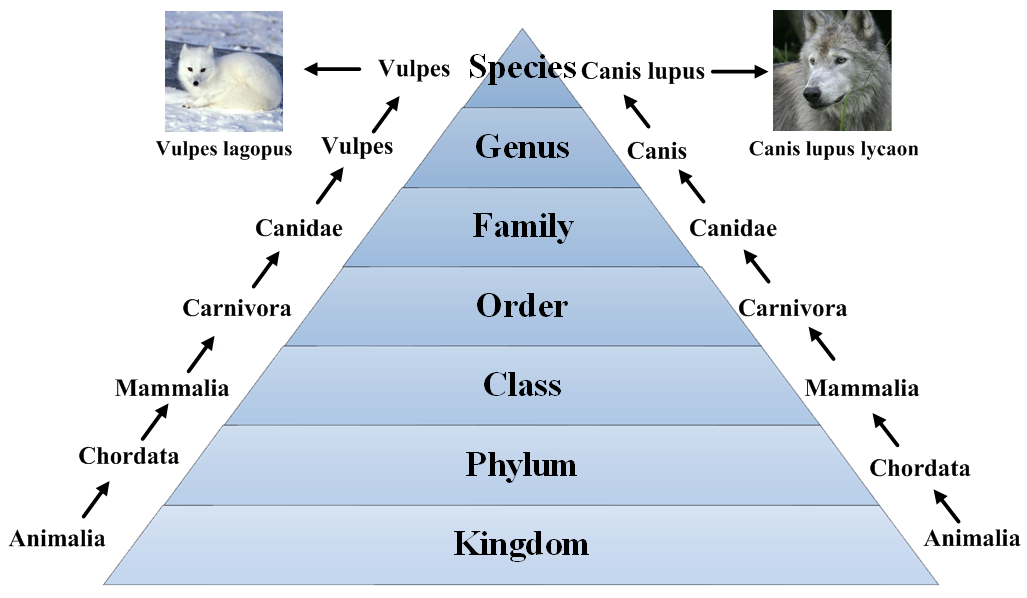}
    \caption{The seven-level classification system of biological taxonomy }\label{bio}
\end{figure}

Biological taxonomy is an important branch of biological research. Its goal is to clarify the kinship between different organisms. The earliest related research can be traced back to 1735 \cite{linnaeus1799species,linnaeus1758systema}. Linnaeus, a Swedish botanist, proposed to divide nature into three realms: plant, animal, and mineral. Plant and animal are further divided into four levels: $Class$, $Order$, $Genus$, and $Species$, thus forming an early classification system. Since then, with the continuous improvement of other biologists, a seven-level classification system of "$Kingdom$-$Phylum$-$Class$-$Order$-$Family$-$Genus$-$Species$" has been formed (as shown in Fig. \ref{bio}). Next, we briefly introduce this classification system.

As shown in Fig. \ref{bio}, the smallest and most basic unit is $species$. If two organisms are belonging to the same species, it means they can share a genetic heritage and produce offspring by mating. For example, two Vulpes Lagopus have the closest relationship because they are the same species.

One level is higher than $species$ is $genus$, which refers to a group of $species$ evolved from a relatively recent common ancestor. For example, Canis Lupus Lycaon and Poodle are animals of the same $genus$ (i.e., Canis), but their kinship is slightly farther than that of the same $species$.

Similarly, the level of $family$ is higher than that of $genus$, and the related $genus$ belong to the same $family$. For example, Vulpes Lagopus and Canis Lupus Lycaon belong to the same $family$ but different $genus$ (as shown in Fig. \ref{bio}).

The $family$ is subordinate to the $order$, the $order$ is subordinate to the $class$, the $class$ is subordinate to the $phylum$, and the $phylum$ is subordinate to the $kingdom$. Correspondingly, as the scope of the concept expands, the kinship gradually decreased.

The closer the kinship, the more common the creatures have, the more similar their characteristics. This law provides us with a solution to measure the degree of correlation of biological image data sets.

\subsection{Multi-task learning}
Modeling some real-life applications can sometimes be difficult to collect enough training data and expensive to accurately label the samples, such as the medical data and rare species data. Multi-task learning (MTL) \cite{ruder2017overview,zhang2017overview} is a technique that can use relevant tasks to assist the decision-making of the model, which can alleviate the problem of data scarcity to some extent. The learning characteristic of MTL is similar to that of human beings, that is, humans can acquire complementary knowledge from related tasks, thereby better solving the current task. MTL also improves the generalization ability of the model by fusing useful information among multiple related tasks.

Taking the neural networks based MTL algorithms as an example, in order to share useful information among tasks, sharing strategies are often used in the network structure and parameter constraints. At present, the commonly used sharing mechanisms include hard parameter sharing \cite{caruana1997multitask} and soft parameter sharing \cite{duong2015low,yang2016trace}. Hard parameter sharing refers to the hidden layers shared by each task and only the last few layers are task-specific. The dual-channel learning framework proposed in this paper is also based on this sharing mechanism (as shown in Fig.~\ref{Multi-task} ). Soft parameter sharing refers to that each task has its own independent structure, but the distance of their parameters are constrained to ensure them to be similar. Here the constraints can be the $l_2$ \cite{duong2015low} distance or the trace norm \cite{yang2016trace}.

One of the difficulties in MTL is the selection of auxiliary tasks. The authors in \cite{caruana1998multitask} believed that the tasks making decisions using the same features are relevant. In \cite{baxter2000model}, the authors pointed out that the tasks with common inductive bias are relevant. In addition, the reference \cite{xue2007multi} mentioned that if the classification boundaries of the two tasks are similar, the two tasks are related. Although these suggestions are helpful for some specific scenarios, they have not been widely used because of the complexity of selection strategies.

\subsection{Zero-shot learning}

ZSL algorithms are mainly used to establish the relationship between the visual space (i.e., the feature vectors extracted from the training data) and the semantic space (i.e., the semantic vectors of the seen and unseen classes). Currently, the training methods of these algorithms can be divided into three categories as follows.

1) Mapping from the visual space to the semantic space. After the model training, the model will be used to classify the semantic space. The typical algorithms include DAP \cite{lampert2013attribute}, ESZSL \cite{romera2015embarrassingly}, and SAE \cite{kodirov2017semantic}.

2) Mapping from the visual space and the semantic space to a third-party embedding space. This method aims to obtain better feature representation in the embedding space and then uses them to make the classification. For example, the CADA-VAE algorithm \cite{schonfeld2019generalized} used in this paper is belonging to this method.

3) Mapping from the semantic space to the visual space. Representative algorithms include DEM \cite{zhang2017learning}, RN \cite{sung2018learning}, CRNet \cite{zhang2019co}, and TCN \cite{jiang2019transferable}.

At present, most of the GZSL algorithms belong to the generation model. As long as the model is able to generate samples of the unseen classes, it can transform ZSL tasks into traditional classification tasks. In addition, this approach can also avoid the problems of hubness \cite{lazaridou2015hubness,shigeto2015ridge} and bias \cite{xian2018feature,zhu2018generative}. For example, CVAE-ZSL \cite{mishra2018generative} generates the unseen classes' samples by learning a Conditional Variational Autoencoder (VAE). f-CLSWGAN \cite{xian2018feature} uses a special classifier to make the features generated by the generator of WGAN more conducive to the final classification.

Considering that GZSL has many advantages and its prediction objectives include both the seen and unseen classes, which is more suitable for real-life application scenarios, the dual-channel learning framework designed in this paper is mainly for GZSL.

%\subsubsection{Generalized zero-shot learning}
\section{The details of the proposed biologically inspired feature enhancement framework}
As mentioned in Sec. I, inspired by the general knowledge of the multi-task learning and biological taxonomy, in this section, we design a novel dual-channel learning framework for ZSL and propose a biologically inspired auxiliary data set selection method for the framework. We study the impact of the auxiliary data sets with different degrees of correlation to the current task on the performance of the ZSL model.

\subsection{The biologically inspired auxiliary data set selection method}
Given a ZSL task, according to the names of the seen classes in its training data set, especially the biological classes, we select three auxiliary data sets with different correlation degrees from ImageNet based on the knowledge of biological taxonomy (i.e., the seven-level classification system shown in Fig. \ref{bio}). For example, if one of the seen classes is dogs, we will select the biological images with different kinship levels to dogs (e.g., same $species$, same $genus$, and same $class$) from ImageNet to form the three auxiliary data sets. For the fairness of the experiment, the number of samples in the three auxiliary data sets is set to be the same. The details of our experimental settings are given in Sec.~\ref{auxilary Datasets}.

\subsection{The details of the dual-channel learning framework}
\subsubsection{Framework}
Our proposed dual-channel learning framework is shown in Fig.~\ref{Multi-task}. In our framework, one channel is trained based on the auxiliary data set and the other channel is trained based on the data set of the current ZSL task (only composed of the training samples from the seen classes). The two channels work together to fine-tune the feature extractor. All modules in the framework (i.e., the feature extractor, the auxiliary task classifier, and the current classifier) are trained together and the optimization objective is as follows:

\begin{figure*}[htbp]
\centerline{\includegraphics[width=14.4cm,height=7.2cm]{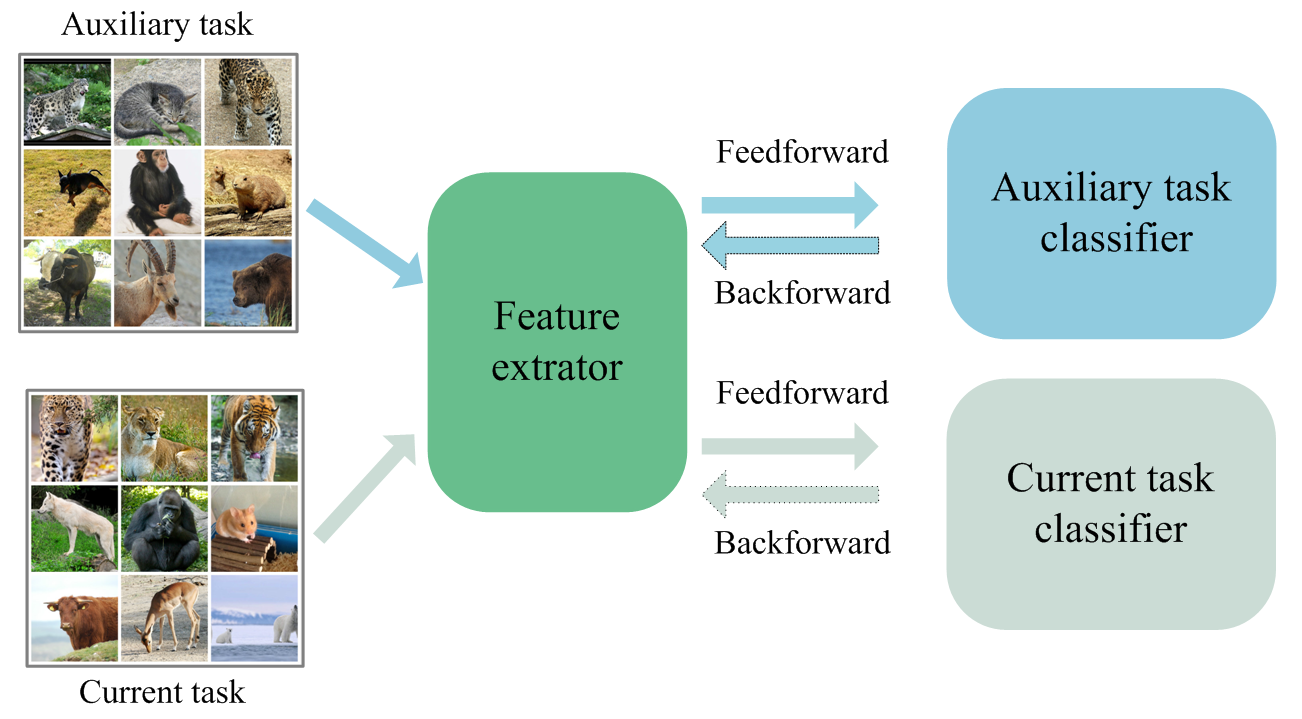}}
\caption{The proposed dual-channel learning framework }
\label{Multi-task}
\end{figure*}

\begin{equation}\label{equ1}
\min_{\theta_{f},\theta_{a}, \theta_{c}} \lambda\mathcal{L}_{aux}+\mathcal{L}_{cur}
\end{equation}
where $\theta_{f}$, $\theta_{a}$, and $\theta_{c}$ are the parameters of the feature extractor, the auxiliary task classifier, and the current classifier, respectively. The goal of model training is to minimize both the loss on the auxiliary task (i.e., $\mathcal{L}_{aux}$) and the loss on the current task (i.e., $\mathcal{L}_{cur}$). Here $\lambda$ refers to the trade-off factor between the auxiliary task and the current task.

\subsubsection{Modules}
\textbf{Feature extractor:} The feature extractor $F$ used in our experiment is ResNet101 \cite{he2016deep} and the parameter of $F$ is marked as $\theta_{f}$. Noted that other deep neural networks such as VGG \cite{simonyan2014very} and Inception \cite{szegedy2015going} can also be used here.

\textbf{Auxiliary task classifier:} The parameter of the auxiliary task classifier $A$ is marked as $\theta_{a}$. $A$ is used to provide additional back-propagation gradients for $F$ by using the auxiliary tasks. Using auxiliary tasks with different correlations to the current task will have different effects on the features extracted by $F$. Most of the traditional classification algorithms such as SVM \cite{cortes1995support} and KNN \cite{cover1967nearest} can be used as $A$.

\textbf{Current task classifier:} The parameter of the current task classifier $C$ is marked as $\theta_{c}$. $C$ is used to fine-tune the parameters of the feature extractor by using the seen classes' training samples to make the extracted features more favorable to the current ZSL task.

\subsubsection{The training mechanism}
As shown in Fig.~\ref{Multi-task}, the feature extractor $F$, the auxiliary task classifier $A$, and the current task classifier $C$ are trained together to minimize Equ. \ref{equ1}. Here we use Equ. \ref{equ2} to represent the output of $F$
\begin{equation}\label{equ2}
  \nu=f_F(x)
\end{equation}

And then the $\mathcal{L}_{aux}$ and the $\mathcal{L}_{cur}$ can be expressed by

\begin{equation}\label{equ3}
  \mathcal{L}_{aux}=l(f_A(\nu_{aux}), y_{aux}
)
\end{equation}
\begin{equation}\label{equ4}
  \mathcal{L}_{cur}=l(f_C(\nu_{cur}), y_{cur})
\end{equation}
where $\nu_{aux}$, $y_{aux}$, $\nu_{cur}$, $y_{cur}$ refer to the output of the feature extractor for the auxiliary task, the labels of the auxiliary samples, the output of the feature extractor for the current task, and the labels of the samples of the current ZSL task, respectively.

Then one can use the gradient descent method to solve Equ.~\ref{equ1}-~\ref{equ4}
\begin{equation}\label{equ5}
  \theta = \theta - \alpha \nabla (\lambda\mathcal{L}_{aux}+\mathcal{L}_{cur})
\end{equation}

\subsection{Performance evaluation}
In this study, we use CADA-VAE to test the impact of different auxiliary data sets on the ZSL model under the proposed framework.

CADA-VAE obtains domain-agnostic representations by aligning the distribution of the feature vector and the side-information in the VAE latent space. After training, CADA-VAE inputs the feature vectors of the seen classes' samples to its encoder to get the corresponding low-dimensional latent features in the VAE latent space.
For the unseen classes, it inputs the side-information of the samples to obtain the corresponding low-dimensional latent features. With the low-dimensional latent features of the seen and unseen classes in the VAE latent space, one can train a classifier for ZSL.

In the testing phase, one can input the feature vectors of the testing samples to the encoder of CADA-VAE to get the low-dimensional latent features, and then use the classifier to predict the corresponding labels.

\section{Experimental settings and results}
\subsection{Datasets}

\subsubsection{Zero-shot data sets}
Three benchmark ZSL tasks are chosen to test the performance of our proposed method, that is, Animals with Attributes2 (\textbf{AWA2}) \cite{xian2018zero}, Caltech-UCSD Bird-200-2011 (\textbf{CUB}) \cite{wah2011caltech}, and A Pascal-a Yahoo(\textbf{APY}) \cite{farhadi2009describing}.

Specifically, \textbf{AWA2} includes 30475 images of 50 kinds of animals.  Among them, 40 are seen classes and 10 are unseen classes, and each class is represented by an 85-dimension vector.
\textbf{CUB} includes 11788 images of 200 kinds of birds. Among them, 150 are seen classes, 50 are unseen classes, and each class is described by a 312-dimension vector.
\textbf{APY} contains 15339 images, 20 classes from the PASCAL VOC2008 database and 12 classes from the Yahoo database. Each class is represented by a 64-dimension vector. We regard the samples from the PASCAL VOC 2008 database as the seen classes and the samples from the Yahoo database as the unseen classes.
The details of each data set and the division of the seen and unseen classes are shown in Table I.
\begin{table}
  \begin{center}
    \caption{The details of the excremental data sets}
    \begin{tabular}{c c c c}
      \hline
      \textbf{Dataset} & \textbf{CUB} & \textbf{AWA2} & \textbf{APY}\\
      %$\alpha$ & $\beta$ & $\gamma$ \\
      \hline
      The number of the images  &11788 &30475 &15339\\
      \hline
      The number of the attributes &312 &85  &64\\
      \hline
      The number of the seen classes &150 &40  &20\\
      \hline
      The number of the unseen classes &50 &10  &12\\
      \hline
    \end{tabular}
  \end{center}
  \label{zero_splits}
\end{table}

\subsubsection{Auxiliary Data sets}\label{auxilary Datasets}
As mentioned in Sec. III, based on the knowledge of biological taxonomy, we select three types of samples from ImageNet that have very low correlation, moderate correlation, and strong correlation with the seen classes of the current ZSL task respectively, and then use them to construct the corresponding low-relevant, middle-relevant, and high-relevant auxiliary data sets.

It is worth mentioning that most of the pre-trained models used by ZSL are trained with ImageNet, and the samples used to construct the auxiliary data set are also from ImageNet, which means that the auxiliary data sets used in our method are very easy to be obtained. Note that the auxiliary data set can also be obtained from other sources such as the Internet based on our proposed selection strategy. Here we take the auxiliary samples directly from ImageNet. The advantage of this is that it is easier to implement. At the same time, we can verify whether we can improve the performance of the ZSL model based on our proposed framework without adding new data sources.

Specifically, given a ZSL task, we choose the low-relevant auxiliary samples from ImageNet based on the filtering criteria of $kingdom$ and then use them to construct the corresponding low-relevant auxiliary data set. Similarly, we construct the middle-relevant auxiliary data set with "the same $kingdom$ but different $class$" as the filtering criteria and the high-relevant auxiliary data set with "the same $class$" as the filtering criteria.

For example, in AWA2, all the seen classes are mammals. In this case, we choose some non-biological images from ImageNet such as water bottles and napkins to construct the low-relevant auxiliary data set. The samples in the middle-relevant auxiliary data set are non-mammal animals such as geckos and tortoises and the samples in the high-relevant auxiliary data set are mammals such as cats. Each auxiliary data set contains 50 classes and each class has the same number of samples.

\subsection{Baseline}
In our experiment, the ResNet-101 pre-trained on ImageNet was chosen as the baseline model. For each ZSL task, we only use the training samples of its seen classes to fine-tune the pre-trained model.

\subsection{Experimental settings}
The original feature extractor used in our framework is the ResNet-101 without its classification layer, and both the auxiliary task classifier and the current task classifier are single-layer network structures. We use Stochastic Gradient Descent (SGD) as the optimizer and the learning rate is set to 0.001. We use the most commonly used Proposed Split (PS) \cite{xian2017zero} as the data division method of CADA-VAE. Other parameter settings of the CADA-VAE algorithm are the same as \cite{schonfeld2019generalized}.

In our experiment, the testing samples can come from both the seen and unseen classes. Suppose $As$ and $Au$ represents the average prediction accuracy of the model on each seen and unseen classes, respectively. Their harmonic mean $\mathcal{H}$ can be expressed as follows.
\begin{equation}\label{equ6}
  \mathcal{H}=\frac{2\times A_s \times A_u}{A_s + A_u}
\end{equation}

At present, $\mathcal{H}$ has become one of the most important indexes to measure the performance of GZSL algorithms. In this paper, we also use $\mathcal{H}$ as the evaluation criterion.

\subsection{Experimental results}

The experimental results on three benchmark data sets are shown in Tabel II.

\begin{table*}[!htp]
  \begin{center}
    \caption{Results of Generalized Zero-shot learning on CUB,AwA2, APY.}
    \begin{tabular}{|c|c|c|c|c|c|c|c|c|c|}
        \hline
        \multirow{2}{*}{Method}&
        \multicolumn{3}{c|}{CUB}&\multicolumn{3}{c|}{AWA2}&\multicolumn{3}{c|}{APY}\cr\cline{2-10}
        &$As$&$Au$&$\mathcal{H}$&$As$&$Au$&$\mathcal{H}$&$As$&$Au$&$\mathcal{H}$\cr
        \hline
        CMT\cite{socher2013zero}&49.8&7.2&12.6 &90.0&0.5&1.0 &85.2 & 1.4&2.8\cr\hline
        SJE\cite{akata2015evaluation}&59.2&23.5&33.6 &73.9&8.0&14.4 &55.7 & 3.7&6.9\cr\hline
        LATEM\cite{xian2016latent} &57.3&15.2&24.0 &77.3&11.5&20.0 &73.0 & 0.1&0.2\cr\hline
        ALE\cite{akata2013label}&62.8&23.7&34.4 &81.8&14.0&23.9 &73.7 & 4.6&8.7\cr\hline
        GAZSL\cite{zhu2018generative}&61.3&31.7&41.8 &86.9&35.4&50.3 &78.6 & 14.2&24.0\cr\hline
        f-CLSWGAN\cite{xian2018feature}&57.7&43.7&49.7 &68.9&52.1&59.4 & - & - &-\cr\hline
        TCN\cite{jiang2019transferable} &52.0&52.6 &52.3 &65.8&61.2&63.4&64.0 & 24.1&35.1\cr\hline
        CADA-VAE\cite{schonfeld2019generalized}&53.5&51.6&52.4&75.0&55.8&63.9 &- & -&-\cr\hline
        CADA-VAE(fine-tuning)&68.4&58.5&63.1& 83.4& 52.0&  64.1&50.0 & 31.8&38.9\cr\hline
        CADA-VAE(low-relevant)&65.0&59.9&62.3&81.4&55.1&65.7 &47.8 & 31.7&38.1\cr\hline
        CADA-VAE(middle-relevant)&68.4&60.6&64.3&84.2&54.3&66.0 &52.9 & 30.2&38.4\cr\hline
        CADA-VAE(high-relevant)&{64.0}&{65.2}&{\bf 64.6}&{79.8}&{57.9}&{\bf 67.1}&{54.3}&{32.3}&{\bf40.5}\cr
        \hline

    \end{tabular}

    \footnotesize{\textbf{Note: } $As$ and $Au$ refer to the accuracy of the model on the seen classes and the unseen classes. $\mathcal{H}$ refers to the harmonic mean of them}
  \end{center}
  \label{zeroshot}
\end{table*}

From Table II, one can observe that the performance of the ZSL model becomes better with the increasing correlation between the auxiliary task and the current ZSL task.
For example, the hybrid accuracy (i.e., $\mathcal{H}$) of the model obtained by training with the low-relevant, middle-relevant and high-relevant auxiliary data sets on AWA2 are 65.7\%, 66.0\%, and 67.1\%, respectively.

\textbf{CADA-VAE (high-relevant) vs Baseline.} Compared with the baseline model (i.e., CADA-VAE-fine-tuning), our method (using the high-relevant auxiliary data sets) can achieve higher prediction accuracy. The accuracy improvement rates on the benchmark data sets CUB, AWA2, and APY are 2.4\%, 4.7\%, and 4.1\%, respectively.
\begin{equation}\label{equ7}
  (64.6\% - 63.1\%)/63.1\% * 100\% = 2.4\%
\end{equation}
\begin{equation}\label{equ8}
  (67.1\% - 64.1\%)/64.1\% * 100\% = 4.7\%
\end{equation}
\begin{equation}\label{equ9}
  (40.5\% - 38.9\%)/38.9\% * 100\% = 4.1\%
\end{equation}
\textbf{CADA-VAE (high-relevant) vs CADA-VAE.} Compared with the original CADA-VAE, one of the best ZSL algorithms, our method can also achieve higher prediction accuracy. The accuracy improvement rates on the CUB and AWA2 are 23.3\% and 5.0\%, respectively.
\begin{equation}\label{equ10}
  (64.6\% - 52.4\%)/52.4\% * 100\% = 23.3\%
\end{equation}
\begin{equation}\label{equ11}
  (67.1\% - 63.9\%)/63.9\% * 100\% = 5.0\%
\end{equation}

In conclusion, the dual-channel learning framework proposed in this paper can effectively improve the generalization ability of the ZSL model with the help of appropriate auxiliary data sets. Specifically, with the help of the auxiliary data sets that are highly related to the current ZSL task, our algorithm has achieved state-of-the-art results in all the three benchmark data sets.

\subsection{An explanation for our experimental phenomena}
\begin{figure*}
\caption{The features of three classes of AWA2 after the dimension reduction}
\centerline{\includegraphics[width=17.4cm,height=5.4cm]{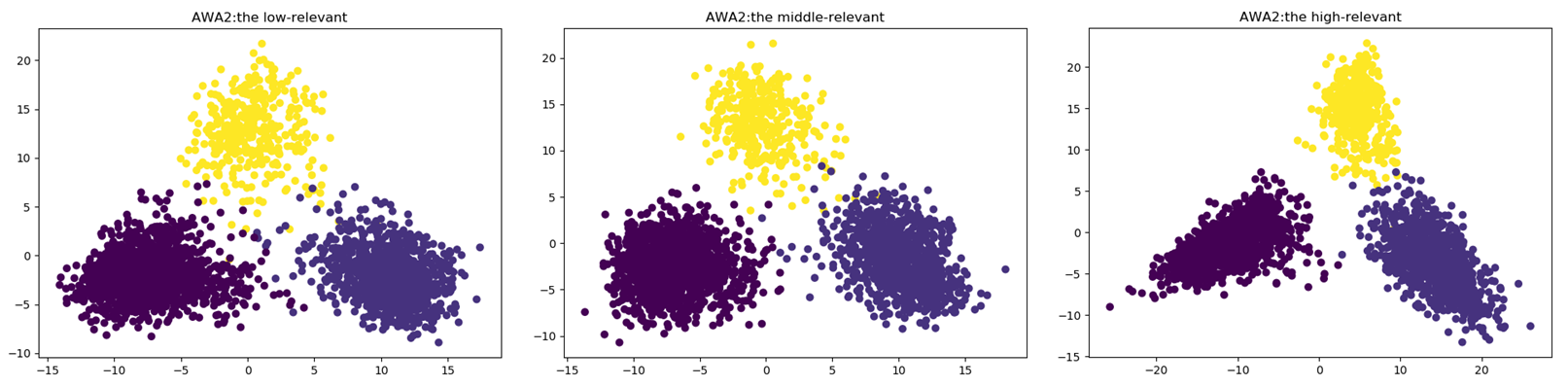}}
\label{visualization}
\end{figure*}

Here we explain the experimental phenomenon from the perspective of feature visualization. Specifically, we analyze the influence of different auxiliary data sets on the final extracted features by using the dimension reduction and visualization method. Due to the experimental results are similar on the three ZSL tasks, so here we take the results on AWA2 as an example (as shown in Fig. \ref{visualization}) to explain the experimental phenomenon. In Fig. \ref{visualization}, each color represents the shape of the features of the samples belonging to a specific class after the dimension reduction.

From Fig. \ref{visualization}, one can observe that as the degree of the correlation between the auxiliary task and the current ZSL task increases, the data features obtained by the ZSL feature extractor show the following rules:  the features of the samples belonging to the same class become more clustered and the features of the samples belonging to different classes become more discrete.

For machine learning, such feature changes are very conducive to the correct decision-making of the final classifier, because the samples belonging to the same class become more clustered and the samples belonging to different classes become more discrete will make the classification easier. Fig. \ref{visualization} also explains the experimental phenomena of this paper to some extent, that is, with the help of the proposed dual-channel learning framework and the auxiliary data sets, the data features of the ZSL task become more separable, so the performance of the final ZSL model is better.

\section {Conclusion}
In this paper, we design a novel dual-channel learning framework for ZSL and propose a new guideline to select auxiliary data sets for the learning framework based on the knowledge of biological taxonomy. Specifically, one can measure the correlation degree of image samples according to the seven-level classification system of the biological taxonomy. We propose to choose the samples closest to the kinship of the seen classes in the current ZSL task to construct the auxiliary data set and then use it to enhance the feature extractor based on our proposed framework. The experimental results on three benchmark data sets (i.e., CUB, AWA2, and APY) show that under the proposed framework, the performance of the ZSL model is gradually improved with the improvement of the correlation degree between the auxiliary data sets and the current ZSL task. This phenomenon is expected to provide a new direction for the future research of ZSL. It is worth mentioning that our algorithm has achieved state-of-the-art results on all three data sets. In the future, we will study to explain the experimental phenomena of this paper mathematically.

%\section*{Acknowledgment}

%The preferred spelling of the word ``acknowledgment'' in America is without
%an ``e'' after the ``g''. Avoid the stilted expression ``one of us (R. B.
%G.) thanks $\ldots$''. Instead, try ``R. B. G. thanks$\ldots$''. Put sponsor
%acknowledgments in the unnumbered footnote on the first page.
\bibliographystyle{siam}

\bibliography{references}

\end{document}